%% file: main.tex
\documentclass{article}

\usepackage[final]{corl_2026} % Uncomment for the camera-ready ``final'' version.
\input{preamble}
\PassOptionsToPackage{numbers,sort&compress}{natbib}
\usepackage[normalem]{ulem}

\title{An Empirical Study of Data-Loading Bottlenecks (and Non-Bottlenecks) in Robotic Imitation Learning}

\author{
  Mrinal Jain$^{1}$ \qquad Arpit Gupta$^{2}$ \qquad Animesh Garg$^{1}$ \\
  $^{1}$Georgia Institute of Technology \\
  $^{2}$General Motors \\
  \texttt{\{mrinal, animesh.garg\}@gatech.edu}, \ \texttt{arpit.gupta@gm.com}
}

\begin{document}
\maketitle

%===============================================================================

\begin{abstract}
    Modern robotic imitation learning pipelines increasingly rely on large-scale datasets, prompting heavy engineering investments in high-performance data infrastructure under the assumption that dataset loading and online augmentation present primary bottlenecks. We challenge this assumption through a systematic, measurement-driven profile of end-to-end policy training within the LeRobot ecosystem across various spatial resolutions. We evaluate standard implementations against a custom, high-performance Rust data engine utilizing Apache Arrow for zero-copy memory access and a copy-on-write (CoW) shared cache. Our profiling reveals three key findings: first, across all reachable configurations on consumer hardware, the data path accounts for a mere 1.0--3.5\% of per-step training time; second, substantial data-path optimizations yielding massive isolated micro-benchmark speedups translate to zero measurable wall-clock improvement in full training; and third, severe data-path degradation stems solely from format-specific serialization flaws, such as unindexed sequential video decoding. Finally, we argue that future robotics data infrastructure must be evaluated against overall training step ratios rather than isolated benchmarks, shifting the community's optimization paradigm away from data loading and toward upstream model layers like visual feature caching. The data engine and training, benchmarking, and evaluation scripts are available at \url{https://github.com/PAIR-data-platform/data-core-rs}.
\end{abstract}


%===============================================================================

\section{Introduction}
Data-driven paradigms have become central to robotic manipulation, leveraging large-scale datasets to train policies via imitation learning (IL) or reinforcement learning (RL)~\citep{act,diffusion_policy,pi0,pi0.5,smolvla,mujoco,pybullet,isaac_sim}. Because high-quality robotics data remains scarce, extensive research focuses on data augmentation tools to artificially expand training distributions. Modern data pipelines involve complex multi-modal ingestion, on-the-fly video decoding, and CPU-bound spatial transformations. Substantial engineering effort targets this layer: general-purpose loaders such as FFCV~\citep{leclerc2023ffcv}, NVIDIA DALI~\citep{nvidia_dali}, and WebDataset~\citep{webdataset} optimize loader throughput; robotics-specific formats such as RoboDM~\citep{robodm} and LeRobot v3~\citep{lerobot} redesign storage and decoding for robot trajectories; and LeRobot itself migrated its video backend from PyAV to torchcodec~\citep{torchcodec} to accelerate frame access (Section~\ref{sec:f3}). This investment presumes that data-path optimizations pay off in end-to-end training time. That premise is what we measure.

To this end, we engineered a high-performance data engine with a Rust backend and Apache Arrow for zero-copy memory access and lazy loading. While isolated micro-benchmarks showed a $\sim$1000$\times$ reduction in single-frame access latency ($\sim$0.8\,$\mu$s from our memory-mapped Arrow buffer vs.\ $\sim$800\,$\mu$s for unindexed disk reads), integrating this engine into end-to-end training loops revealed a counterintuitive reality: faster data loaders yield no measurable wall-clock improvement in practice. Profiling standard configurations within the LeRobot ecosystem shows that the data path consumes a negligible 1.0--3.5\% of per-step training time; the remaining budget is entirely dominated by the forward and backward passes of the deep neural network models. Per Amdahl's law, isolated data loading speedups completely evaporate during full-lifecycle training.

This paper provides a measurement-driven profile of modern robot policy training. We detail our systems architecture, document the engineering constraints of per-worker caching (such as Out-Of-Memory ceilings), and introduce a copy-on-write (CoW) shared cache framework to mitigate memory overhead. While we contribute optimized online augmentation pipelines directly to the LeRobot infrastructure, our primary contribution is paradigm-shifting: we argue that the robot learning community is actively optimizing the wrong layer. Future infrastructure work must justify its utility against the actual data-loading-to-model-update ratio under target training configurations, redirecting structural engineering efforts away from data-loading abstractions and toward upstream computational bottlenecks.

%===============================================================================

\section{Background and Related Work}

% \subsection{Supervised Learning Approaches in Robotics}
\paragraph{Supervised Learning Approaches in Robotics}
Supervised policy learning has transitioned from low-dimensional state control to high-frequency, multi-modal behavior cloning, scaling both model capacity and data throughput demands. For instance, the Action Chunking Transformer (ACT)~\citep{act} utilizes a conditional variational autoencoder requiring synchronized multi-view frames and historical context, making random-access reads to temporal video the dominant I/O pattern. Conversely, Diffusion Policy~\citep{diffusion_policy} formulates action generation as an iterative denoising process, shifting the computational weight to the neural optimizer. Vision-Language-Action (VLA) models (e.g., $\pi_{0.5}$~\citep{pi0.5}, SmolVLA~\citep{smolvla}, DreamVLA~\citep{dreamvla}) introduce heterogeneous multi-modal inputs but maintain structurally identical temporal data streaming paths.

% \subsection{ML Data Infrastructure}
\paragraph{ML Data Infrastructure}
An ecosystem of data-loading libraries targets throughput demands in large-scale ML. Tools like \texttt{PyAV}, \texttt{decord}, and \texttt{torchcodec} handle video decoding, while frameworks like FFCV~\citep{leclerc2023ffcv}, WebDataset, and NVIDIA DALI optimize sequential throughput via memory-mapping, sharding, or GPU-side preprocessing. However, these libraries are almost exclusively evaluated using isolated benchmarks (e.g., samples per second) rather than as a fraction of total training step time under realistic prefetching. This evaluation methodology misleads when model compute heavily dominates step time.

% \subsection{Data Augmentation in Robot Imitation Learning}
\paragraph{Data Augmentation in Robot Imitation Learning}
We characterize augmentation into two categories:
\begin{itemize}
    \item \textbf{One-to-one transformations:} Pixel-level perturbations (e.g., color jitter, random cropping) or sensor noise are applied to trajectories independently on CPU workers. Though computationally simple and discarded after use, they significantly impact policy performance~\citep{young2021visual,kostrikov2020image}.
    \item \textbf{Many-to-many transformations:} Methods like MimicGen, CACTI, or GenAug~\citep{mimicgen,dexmimicgen,scaling_imagened,cacti,rocoda,genaug} generate new trajectories via simulation or apply spatially consistent perturbations across entire sequence windows. Because they impose synchronized dependencies across multi-camera frames, they motivate data engines capable of structuring multi-modal windows before tensor materialization. This requirement drives our window-structured intermediate representation (Section~\ref{sec:engine}), over which one-to-one operations are then fused (Section~\ref{sec:f4}).
\end{itemize}

% \subsection{Robot IL Data Formats and the LeRobot Ecosystem}
\paragraph{Robot IL Data Formats and the LeRobot Ecosystem}
Robotics storage formats have evolved from \texttt{rosbag} (which lacks efficient random-access for training) to structured layouts like HDF5~\citep{hdf5} and Zarr for index-based slice reads. To support expanding dataset scales~\citep{droid,open_x_embodiment}, contemporary formats like RoboDM~\citep{robodm} and LeRobot v3~\citep{lerobot} store visual data as AV1-compressed video, reconstructing pixel arrays on the fly using background threads. This delta-timestamp pattern minimizes storage footprints but requires complex temporal indexing and sequential seeks.

% \subsection{Systems Profiling and Bottleneck Attribution}
\paragraph{Systems Profiling and Bottleneck Attribution}
Attributing end-to-end training time to specific pipeline stages is well-established in systems literature but largely absent from robot learning~\citep{zhang2019empirical}. While computer vision studies highlight data-loading bottlenecks in large-batch image classification~\citep{mohan2020analyzing}, these findings do not translate directly to robot IL, which features expensive multi-camera video decoding alongside large transformer architectures with high compute-to-data ratios. This work fills this gap by providing systematic, step-level profiling.

%===============================================================================

\section{Methods}
\subsection{The Training Step and Data Path Metric}
We profile policy training directly within the active LeRobot training loop. Each optimization step decomposes into two concurrent paths: (1) the \emph{data path}, which handles frame decoding, tensor materialization, online augmentation, and host-to-device PCIe transfer; and (2) the \emph{compute path}, which executes the forward pass, loss computation, backward pass, and optimizer update.

Standard data loaders overlap these paths via prefetching: while the accelerator executes step $t$, background workers prepare batch $t+1$. Consequently, the data path impacts wall-clock time only when it cannot be hidden behind compute. We instrument the loop to log two per-step quantities: \texttt{data\_s} (the time the trainer blocks waiting for a batch) and \texttt{updt\_s} (the execution time of the compute path).

Our primary metric is the \emph{data path ratio} $r$:
$$r = \frac{\texttt{data\_s}}{\texttt{updt\_s}}$$
For small values, $r$ approximates the fraction of step time the GPU spends stalled. We use this ratio rather than $\texttt{data\_s}/(\texttt{data\_s}+\texttt{updt\_s})$ because in pathological cases where data loading stalls exceed an entire compute step, $r$ remains easily legible as $r > 100\%$.

Crucially, \texttt{data\_s} is a \emph{residual cost}, not an absolute measure of decode compute. A heavy decode pipeline may yield $\texttt{data\_s} \approx 0$ if fully hidden behind a slow model step, while an underperforming decoder can yield high \texttt{data\_s} even with lightweight models. Thus, while \texttt{data\_s} precisely attributes wall-clock training overhead, isolated benchmarks are required to capture true raw decode costs.

\subsection{The Data Engine}
\label{sec:engine}
\begin{figure}[t]
    \centering
    \includegraphics[width=0.95\linewidth]{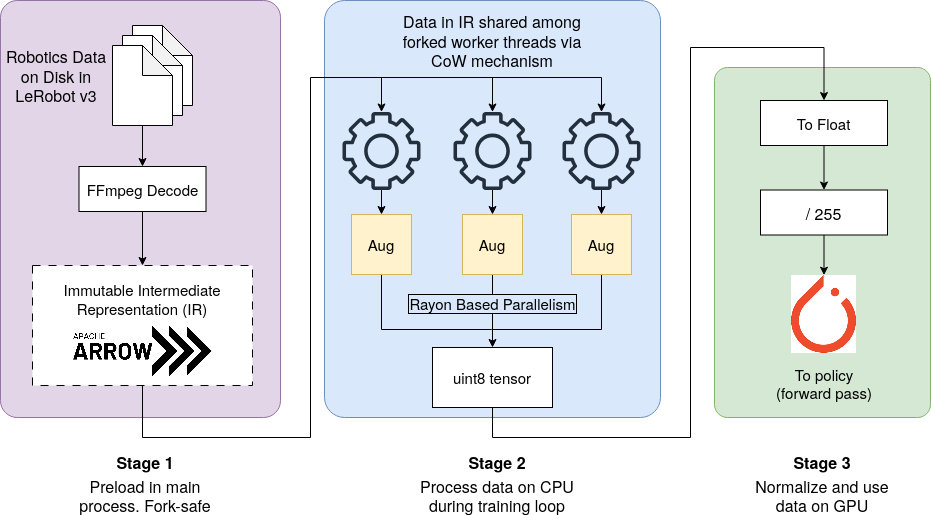}
    \caption{Example data path with our high performance backend using the \texttt{arrow\_uint8} strategy. Data is preloaded into an Intermediate Representation (Stage 1), worker processes are forked (Stage 2), and finally data is normalized on the GPU (Stage 3).}
    \label{fig:data_path}
\end{figure}

To establish a high-performance ceiling, we implement a data engine in Rust and evaluate progressively optimized designs against standard torchcodec-based dataloaders.

\textbf{Zero-copy Apache Arrow Intermediate Representation (IR).} Central to our engine is decoding video episodes once into an immutable Apache Arrow columnar buffer. Frame access is reduced to an $O(1)$ pointer lookup, allowing bytes to cross into PyTorch without serialization or copy operations.

\textbf{CoW Cache Sharing.} Rather than maintaining separate per-worker caches (which scales memory requirements linearly and duplicates decode work) our final design pre-decodes episodes in the parent process. Forked workers inherit the immutable Arrow pixel buffer via the Linux Copy-on-Write (CoW) mechanism. Because workers only execute reads, they share a single virtual memory mapping, eliminating both redundant decodes and inter-process serialization overhead. This centralized, synchronous preloading strategy also bypasses critical library concurrency constraints associated with multi-threaded decoding backends (see Appendix for a detailed discussion).

\textbf{Materialization Strategies.} We evaluate three strategies for transferring frames across the Rust-Python-GPU boundary, trading off host memory, PCIe bandwidth, and GPU compute:
\begin{itemize}
    \item \texttt{arrow\_f32}: The HWC-to-CHW transpose, \texttt{uint8}-to-\texttt{float32} cast, and $/255$ normalization occur in Rust via the Rayon library. This minimizes GPU compute but quadruples the pin-memory copy and PCIe host-to-device transfer volume due to floats being $4\times$ larger than \texttt{uint8}.
    \item \texttt{arrow\_uint8}: Rust returns raw HWC \texttt{uint8} bytes. Tensors cross the PCIe bus as integers, and float normalization is executed directly on the GPU, reducing bus traffic at the expense of overlapping normalization with model compute.
    \item \texttt{arrow\_uint8\_jpeg}: The cache retains compressed JPEG bytes, which are decoded to raw integer pixels in Rust via Rayon on every access before following the downstream path of \texttt{arrow\_uint8}. This minimizes the memory footprint of the CoW cache at the cost of repeated CPU decode overhead.
\end{itemize}

\textbf{Fused Intermediary Augmentations.} One-to-one augmentations (e.g., color jitter, cropping) are applied directly to integer RGB pixels before tensor materialization in our pipeline. These operations are fused into a single parallelized Rayon pass, guaranteeing temporal consistency across frames within a given temporal window.

%===============================================================================

\section{Results}
\label{sec:results}

We profile the LeRobot training data path across three datasets, three spatial resolutions, four batch sizes, two policy classes (ACT~\citep{act} and SmolVLA~\citep{smolvla}), and two accelerators: a consumer RTX~4070~Ti~Super and a server-class L40S. We vary only the data backend and augmentation to isolate data-path effects. All experiments use 4 dataloading workers and a prefetch factor of~2. Full hardware and configuration details are in the Appendix. To measure GPU metrics such as GPU utilization, we use the NVIDIA Management Library (NVML) sampling at 20\,ms intervals.

% =====================================================================
\subsection{Data Loading is Hidden by Model Compute}
\label{sec:f1}

\begin{figure}[t]
  \centering
  \includegraphics[width=0.98\linewidth]{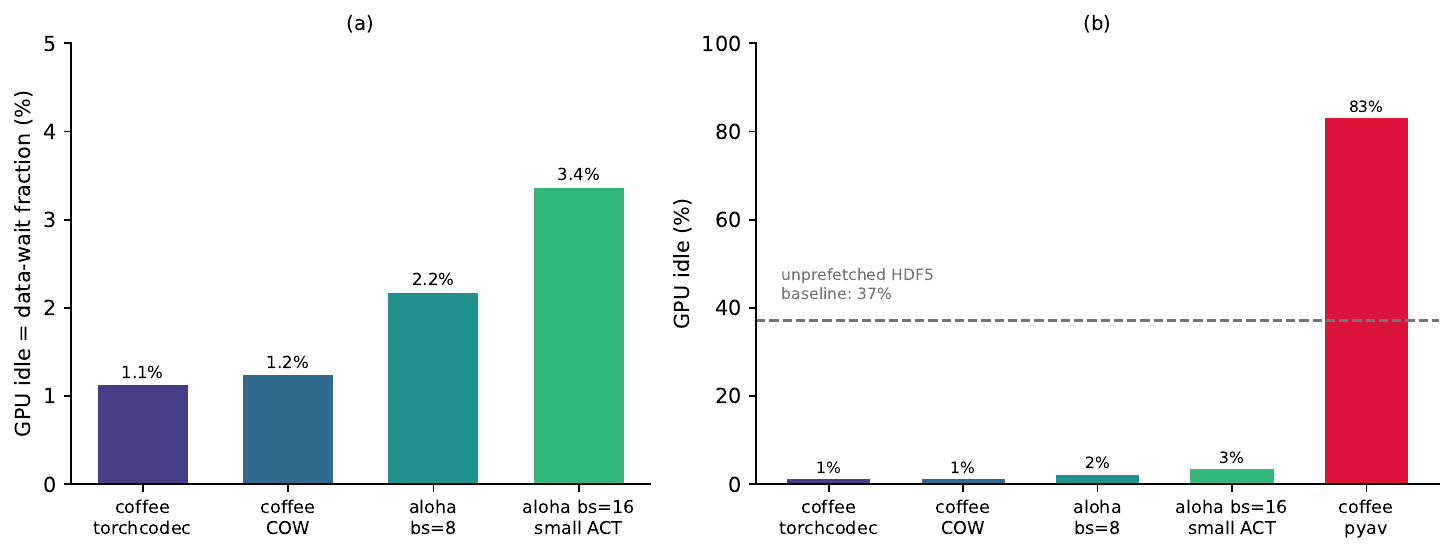}
  \caption{Per-step GPU-idle (data-wait) fraction. (a) Standard configurations
  exhibit minimal data-starvation (1.1--3.4\%). (b) Pathological PyAV
  sequential decoding inflates the idle fraction to $\sim$83\%; the dashed line
  indicates a baseline unprefetched HDF5 implementation (37\% idle).}
  \label{fig:f1}
\end{figure}

Across all tested resolutions ($84\times84$ to $480\times640$), batch sizes (8--64), and backends, the data-path ratio $r$ stays within a narrow 1.0--3.5\% band (Table~\ref{tab:ratio}, Figure~\ref{fig:f1}). Model compute dominates step time (168--368\,ms for forward/backward passes), leaving unhidden loader wait times of just 2--13\,ms. Notably, the ratio is insensitive to data volume: raising resolution or batch size does not push $r$ above 3.5\%. The only lever that meaningfully increases $r$ is \emph{reducing model capacity}---the smaller-ACT variant at bs=16 reaches 3.5\%, the highest standard value we observe. The full per-window distribution (Appendix~\ref{app:supp}) confirms that every configuration sits below 5\%. We use 5\% as a practical heuristic rather than a derived constant: under Amdahl's law, $r$ bounds the largest wall-clock gain any loader improvement can deliver, and below $\sim$5\% that bound is comparable to run-to-run variance in end-to-end training time (Table~\ref{tab:wallclock}).

\begin{table}[t]
  \centering
  \caption{Steady state per-step timing of data path and compute path latency (mean $\pm$ std over 100-step windows). The ratio $r$ is calculated with the means.}
  \label{tab:ratio}
  \begin{tabular}{lrrr}
    \toprule
    Config & \texttt{data\_s} (ms) & \texttt{updt\_s} (ms) & Data-Path Ratio $r$ \\
    \midrule
    coffee $84\times84$, bs=64 (torchcodec)   & $1.9 \pm 1.5$  & $167.6 \pm 0.8$ & {1.13\%} \\
    coffee $84\times84$, bs=64 (CoW)          & $2.1 \pm 1.1$  & $167.6 \pm 0.6$ & 1.23\% \\
    aloha $480\times640$, bs=8 (torchcodec)   & $6.9 \pm 0.2$  & $311.1 \pm 3.3$ & {2.23\%} \\
    aloha $480\times640$, bs=16 (smaller ACT) & $12.8 \pm 0.4$ & $367.8 \pm 5.9$ & {3.49\%} \\
    coffee $84\times84$, bs=64, \textbf{PyAV} & $\sim$810      & $\sim$166       & \textbf{$\sim$488\%} \\
    \bottomrule
  \end{tabular}
\end{table}

% =====================================================================
\subsection{Isolated Loader Speedups Do Not Map to Training Throughput}
\label{sec:f2}

Because the data path is almost entirely hidden behind model compute, even dramatic isolated improvements evaporate in end-to-end training. Table~\ref{tab:isobench} shows that our \texttt{arrow\_uint8} backend outperforms torchcodec by $3.94\times$ and PyAV by $6.95\times$ in single-process throughput (no DataLoader workers) on a 4-camera $480\times640$ dataset. A warm frame cache is even more extreme: the CoW shared cache serves $\sim$625k samples/s via an $O(1)$ pointer read, a $\sim$$10^6\times$ advantage over cold-cache disk video decoding ($\sim$0.11 samples/s) on the same dataset (Figure~\ref{fig:f2}). Yet in the active training loop, both advantages vanish: across three seeds, end-to-end wall-clock differences between backends are within run-to-run noise (Table~\ref{tab:wallclock}).

The \texttt{arrow\_f32} result in Table~\ref{tab:isobench} illustrates that poorly placed optimizations can actively \emph{harm} throughput: offloading \texttt{uint8}-to-\texttt{float32} normalization to the CPU quadruples the host memory footprint and PCIe transfer volume, cutting isolated throughput nearly in half under heavy augmentation and dropping below the torchcodec baseline. This underscores that data-path design choices must account for the full memory hierarchy, not just decode latency.

\begin{table}[!t]
  \centering
  \caption{Isolated single-process (no DataLoader workers) pipeline decoding and loading throughput (samples/sec). The \texttt{arrow\_uint8} strategy has significant speedup in an isolated benchmark compared to the torchcodec baseline. \texttt{arrow\_f32} fuses normalization into the augmentation pass; after its heavy-aug PCIe regression we discontinued the strategy, so its no-aug cell was not measured.}
  \label{tab:isobench}
  \begin{tabular}{lrrr}
    \toprule
    Backend & no-aug & heavy-aug & vs torchcodec (no-aug) \\
    \midrule
    \texttt{arrow\_uint8} (Ours, $/255$ on GPU) & \textbf{135.5} & 48.1 & \textbf{3.94$\times$} \\
    \texttt{arrow\_f32}   (Ours, $/255$ on CPU) & ---            & 25.5 & --- \\
    torchcodec (baseline)                       & 34.4            & 26.3 & 1.0$\times$ \\
    PyAV                                        & 19.5            & 16.0 & 0.57$\times$ \\
    \bottomrule
  \end{tabular}
\end{table}

\begin{table}[!t]
  \centering
  \caption{End-to-end training-loop wall-clock duration (coffee $84\times84$, bs=64; mean $\pm$ std over seeds 42--44 where marked). Steady-state loop only: excludes the one-time pre-decode ($\sim$50\,s per 100 episodes), which would only make our backend slower. Backend differences are within run-to-run noise.}
  \label{tab:wallclock}
  \begin{tabular}{lrr}
    \toprule
    Backend & training-loop wall-clock & vs torchcodec \\
    \midrule
    torchcodec (baseline, 3 seeds)         & $326 \pm 4$~s & --- \\
    CoW raw-pixel (3 seeds)                & $324 \pm 1$~s & $-0.6\%$ \\
    \texttt{arrow\_uint8\_jpeg} (1 seed)   & 329~s & $+0.9\%$ \\
    \bottomrule
  \end{tabular}
\end{table}

% =====================================================================
\subsection{Video Indexing Failures Induce Data Bottlenecks}
\label{sec:f3}

The PyAV configuration is the sole exception where the data path bounds training; the root cause is a \emph{format flaw}, not a fundamental limitation of data loading. PyAV performs $O(N)$ sequential seeks on AV1-compressed video that lacks frame-level indices. On the coffee dataset this produces \texttt{data\_s} $\approx 0.81$\,s against \texttt{updt\_s} $= 0.166$\,s ($r \approx 488\%$), inflating total training time from 5.8 to 32.2 minutes ($5.6\times$ slower). At batch size 32, the gap widens to $15.1\times$ slower, driving LeRobot's recent switch to torchcodec. We stress that this is a property of the decoder's seek path, not of the LeRobot format: the same AV1 files read through torchcodec's indexed seeking yield $r = 1.1\%$ (Table~\ref{tab:ratio}), and LeRobot's current default loader sits comfortably in the compute-masked regime. The takeaway for practitioners is simple: pair storage formats with a decoder that supports index-based random access, and severe data bottlenecks do not arise.

% =====================================================================
\subsection{Online Augmentation Overhead Remains Negligible}
\label{sec:f4}

Online augmentation adds at most 0.3 percentage points to the data-path ratio (Table~\ref{tab:aug}). Our Intermediate Representation (IR) fused augmentation pipeline---which applies color jitter and cropping to integer pixels before tensor materialization---matches torchvision v2 in end-to-end training, despite outperforming it by $1.83\times$ in isolated throughput under heavy augmentation (48.1 vs 26.3 samples/s). The pattern mirrors Section~\ref{sec:f2}: compute masking erases the isolated advantage entirely.

\begin{table}[t]
  \centering
  \begin{minipage}[c]{0.4\linewidth}
  \centering
  \caption{Data-path ratio $r$ across backend and augmentation configurations (coffee $84\times84$, bs=64). Even with augmentations, the ratio $r$ is still low.}
  \label{tab:aug}
  \end{minipage}
  \begin{minipage}[c]{0.58\linewidth}
  \centering
  \resizebox{0.9\linewidth}{!}{
  \begin{tabular}{lcc}
    \toprule
    Augmentation pathway          & no-aug & +aug \\
    \midrule
    torchcodec (torchvision v2)   & 1.13\% & 1.29\% \\
    CoW (Rust-IR fused)           & 1.23\% & 1.49\% \\
    \bottomrule
  \end{tabular}
  }
  \end{minipage}
\end{table}

% =====================================================================
\subsection{Generalization to Large-Scale VLA Training}
\label{sec:f5}

\begin{figure}[h!]
  \centering
    \begin{minipage}[c]{0.7\linewidth}
  \centering
  \includegraphics[width=\linewidth]{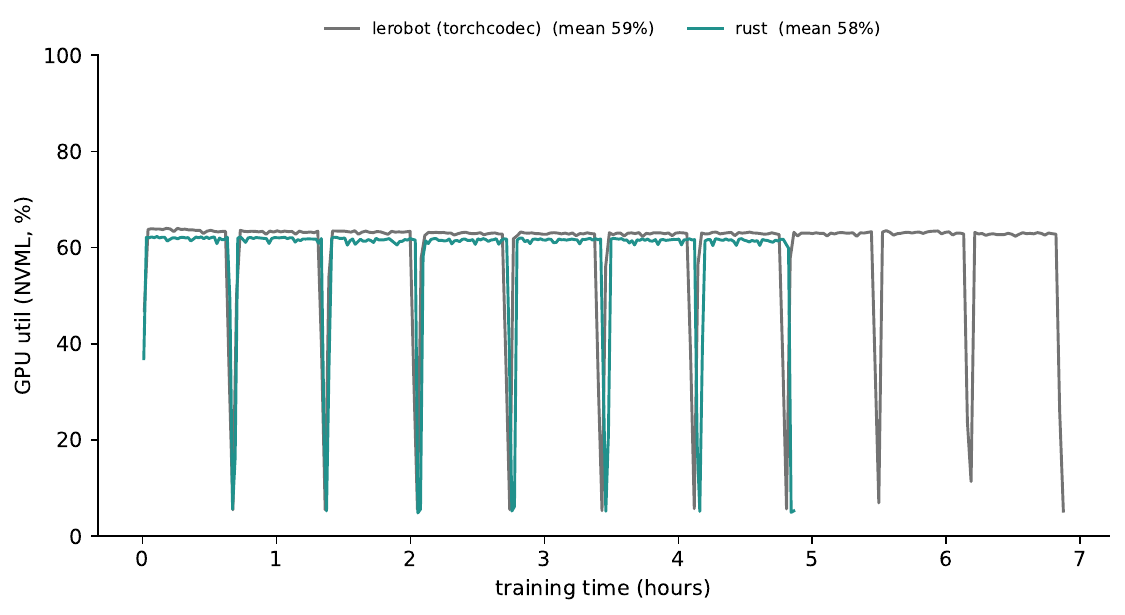}
  \end{minipage}
 \begin{minipage}[c]{0.29\linewidth}
  \centering
  \caption{NVML utilization over a multi-hour run, torchcodec vs Rust overlaid
  (supplements Section \ref{sec:f5}): steady $\sim$59--61\% with periodic downward
  spikes that are eval/checkpoint (sim rollouts), not data stalls.}
  \label{fig:smolvla_ts}
  \end{minipage}
\end{figure}

\begin{figure}[t]
  \centering
  \begin{minipage}[c]{0.7\linewidth}
  \centering
  \includegraphics[width=\linewidth]{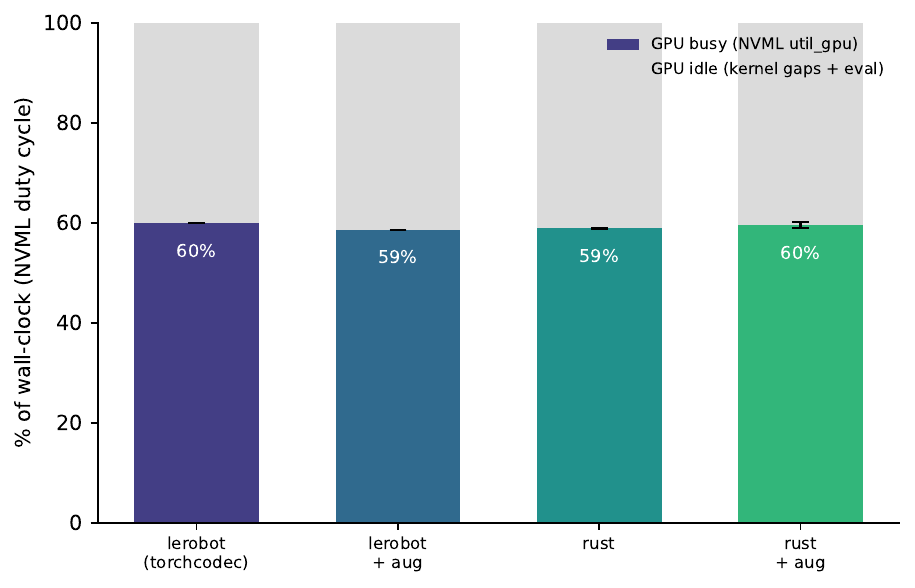}
  \end{minipage}
  \begin{minipage}[c]{0.28\linewidth}
  \centering
  \caption{SmolVLA / LIBERO-10 GPU utilization. The duty cycle remains
  invariant within 1.3 percentage points across backends and augmentations,
  demonstrating that the $\sim$40\% idle time is framework and evaluation
  overhead rather than data starvation.}
  \label{fig:smolvla_bar}
  \end{minipage}
\end{figure}

To test whether compute-masking is an artifact of lightweight ACT on consumer hardware, we profile SmolVLA~\citep{smolvla} on LIBERO-10~\citep{libero} using a server-class L40S GPU. Across all four backend configurations (stock LeRobot (torchcodec), LeRobot with aug, our Rust engine, and Rust engine with aug), GPU utilization is invariant within 1.3 percentage points (59--60\%; Table~\ref{tab:smolvla}, Figure~\ref{fig:smolvla_bar}). The $\sim$40\% idle fraction is attributable to Python host overhead and periodic evaluation rollouts, not data stalls, as confirmed by the multi-hour utilization trajectory (Figure~\ref{fig:smolvla_ts}), where the only drops coincide with checkpointing and rollout phases. Compute-masking of the data path is therefore not hardware- or model-class-specific.

\paragraph{Scaling beyond a single GPU.} Two further regimes follow from the same accounting. Under standard data-parallel training (e.g., PyTorch DDP), each GPU rank runs its own dataloader workers and prefetch queue, so the per-rank ratio $r$ is governed by the same per-step \texttt{data\_s}/\texttt{updt\_s} trade-off measured here; as model scale grows, \texttt{updt\_s} grows faster than \texttt{data\_s}, pushing $r$ further toward zero. With network-attached storage (NFS, object stores), prefetching and the page cache hide I/O latency behind compute in steady state unless storage bandwidth falls below the tensor ingestion rate. Because $r$ requires only the \texttt{data\_s} and \texttt{updt\_s} values LeRobot already logs, practitioners can check either regime directly on their own workloads.

\begin{table}[t]
  \centering
  \begin{minipage}[c]{0.3\linewidth}
  \caption{SmolVLA / LIBERO-10 on L40S: NVML GPU duty cycle across backends.}
  \label{tab:smolvla}
  \end{minipage}
  \begin{minipage}[c]{0.69\linewidth}
  \resizebox{\linewidth}{!}{
  \begin{tabular}{lcc}
    \toprule
    Backend & GPU util (NVML duty cycle) & mem used \\
    \midrule
    lerobot (torchcodec)      & \textbf{60\%} & 11.5 / 48~GB \\
    lerobot + aug             & 59\%          & 11.4 / 48~GB \\
    rust (\texttt{data\_core})& 59\%          & 11.5 / 48~GB \\
    rust + aug                & \textbf{60\%} & 11.5 / 48~GB \\
    \bottomrule
  \end{tabular}
  }
  \end{minipage}
\end{table}

% =====================================================================
\subsection{Feature Caching Targets the Actual Bottleneck: Model Compute}
\label{sec:f6}

Since the data path is not the limiting factor, reducing \texttt{updt\_s} is the productive optimization target. The dominant repeated cost within \texttt{updt\_s} is the per-camera visual backbone forward pass, which runs on every training step even though the same frames are revisited across epochs. Caching frozen encoder outputs eliminates this redundancy.

We validate this on ACT with a frozen ResNet-18 backbone (Table~\ref{tab:feature}). Caching reduces training time by \textbf{1.3--1.5$\times$}: aloha (40 episodes, spatial map features) drops from 89.1 to 61.6\,s and coffee (80 episodes) from 68.7 to 51.9\,s. Switching from spatial map features to pooled features yields an additional \textbf{2.5$\times$} speedup (61.6$\to$24.6\,s) and a \textbf{47.5$\times$} reduction in cached storage (11.05\,GB$\to$232.5\,MB).

In imitation learning, the metric that matters is closed-loop task success, not training loss~\citep{ross2011dagger}. We use L1 for one narrow purpose: certifying that the cache is \emph{faithful}. The same action head trained on cached features and on the identical frozen backbone run online reaches matching train L1 (0.0623 vs.\ 0.0624), ruling out train/inference feature skew. For closed-loop evaluation we compare against a stock LeRobot ACT (the paper architecture, trainable backbone, same data, 10k steps, bs=64) over 25 rollouts from deterministic MimicGen initial states. Both policies score \textbf{0/25}. The stock policy fails identically despite converged training (L1 0.156$\to$0.074), so the failure stems from the short training budget rather than from caching (Appendix~\ref{app:feature_cache}). Feature caching is therefore a validated \emph{systems} lever whose policy-level benefit remains to be established at a training budget where the uncached baseline succeeds. None of our central claims (the 1.0--3.5\% ratio and the Amdahl envelope) depends on it.

\begin{table}[t]
  \centering
  \caption{Feature-caching performance metrics: training speed, storage
  footprints, cache fidelity (L1), and closed-loop success against a stock ACT baseline.}
  \label{tab:feature}
  \resizebox{\linewidth}{!}{
  \begin{tabular}{l p{0.58\linewidth}}
    \toprule
    Comparison & Value \\
    \midrule
    Cache vs frozen-online, aloha 40ep (map) & $89.1 \rightarrow 61.6$~s = \textbf{1.45$\times$}, train L1 0.0623 vs 0.0624 \\
    Cache vs frozen-online, coffee 80ep (map) & $68.7 \rightarrow 51.9$~s = \textbf{1.32$\times$} \\
    Pooled vs map, aloha train time & $24.6$ vs $61.6$~s = \textbf{2.5$\times$} \\
    Pooled vs map, disk & $232.5$~MB vs $11.05$~GB = \textbf{47.5$\times$} \\
    Pooled vs map, held-out L1 (aloha) & \textbf{0.0657 vs 0.0730} (pooled better) \\
    3-seed held-out L1 (coffee map / aloha pooled) & $0.0954 \pm 0.0005$ / $0.0610 \pm 0.004$ \\
    MimicGen rollout success, cached map (coffee, 10k steps) & \textbf{0 / 25} \\
    MimicGen rollout success, stock LeRobot ACT (coffee, 10k steps) & \textbf{0 / 25} \\
    \bottomrule
  \end{tabular}
  }
\end{table}

% =====================================================================
\subsection{The Amdahl Envelope}
\label{sec:amdahl}

\begin{figure}[t]
  \centering
  \includegraphics[width=0.9\linewidth]{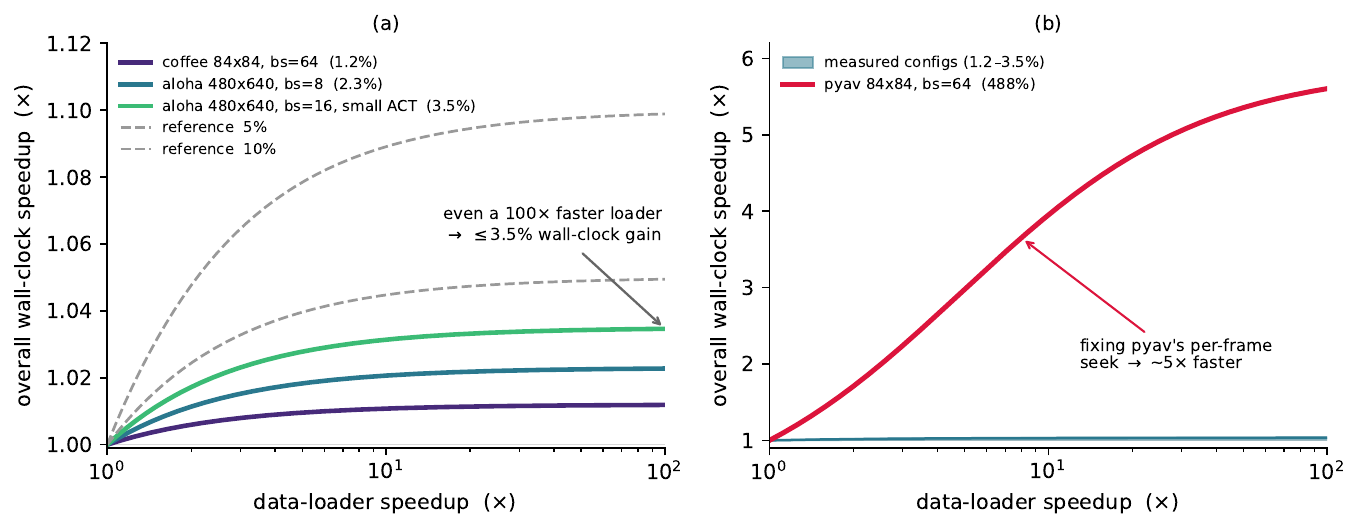}
  \caption{Amdahl scaling envelope for data-path acceleration. (a) At
  empirical ratios (1.0--3.5\%), a $100\times$ loader speedup yields $\leq$3.5\%
  overall wall-clock improvement. (b) Significant scaling headroom
  ($\sim$5.9$\times$) only emerges within the pathological PyAV decoding
  regime.}
  \label{fig:amdahl}
\end{figure}

Figure~\ref{fig:amdahl} makes the cumulative argument precise. At our empirical ratios (1.0--3.5\%), Amdahl's law bounds the maximum possible wall-clock gain from any data-loader improvement to $\leq$3.5\%, regardless of the speedup factor; even an infinitely fast loader cannot do better (Panel~a). The PyAV regime ($r \approx 488\%$) is the instructive counterexample. Fixing the index flaw yields up to $\sim$5.9$\times$ end-to-end improvement (Panel~b). Together, the two panels define a decision rule: data-infrastructure investment is only justified when $r$ is measured to be large under target training conditions.

\paragraph{Explicit Non-Claims.}
To clarify scope: our Rust pipeline is highly competitive in isolation---the issue is that isolation is the wrong evaluation frame. CoW memory-sharing functions as designed but cannot shift end-to-end wall-clock at our operating points. We do not assert that \emph{all} robotic training loops are compute-bound, only that standard single-GPU ACT and VLA setups within the LeRobot community are. PyAV's degradation is specific to random-access seeking in AV1 containers, not a general indictment of the library. And the $\sim$40\% SmolVLA NVML idle fraction is a framework-overhead invariance test, not evidence of data stalls.

%===============================================================================
%===============================================================================

\section{Discussion}
\subsection{Data Loading is Seldom the Bottleneck}
Our empirical profiling within the LeRobot ecosystem challenges the assumption that the data path represents a critical computational bottleneck in supervised robot learning. Across all hardware and policy configurations that we tested, the data path occupies just 1.0--3.5\% of the per-step training budget. Consequently, while our high-performance Rust data engine (built with Apache Arrow and a copy-on-write shared cache) achieved a $\sim$1000$\times$ reduction in single-frame access latency and a $3.94\times$ isolated pipeline speedup, it yielded no wall-clock improvement beyond run-to-run noise in full training loops.

This outcome is strictly governed by Amdahl’s law. Because the compute path completely dominates step time, even an infinitely fast data loader upper bounds overall speedups to 3.5\%. This demonstrates that isolated micro-benchmarks (e.g., samples per second) are highly misleading metrics for robotic data infrastructure. Engineering efforts should shift away from lower-level data-loading abstractions and redirect toward upstream computational bottlenecks.

\subsection{The Severe Cost of Unindexed Sequential Decoding}
When data-path degradation does occur, it is a format-specific serialization flaw rather than a systemic failure of data-loading architecture. The severe performance regression observed with PyAV stems entirely from $O(N)$ unindexed, sequential frame seeking on AV1-compressed video. For systems designers, this implies that infrastructure complexity like custom shared-memory daemons is often unnecessary. Optimal data ingestion baselines can be achieved simply by choosing storage formats that natively support robust, index-based random access.

%===============================================================================
%===============================================================================

\section{Limitations}
\begin{itemize}
    \item \textbf{Scope of hardware setup.} Our evaluation focuses on single-GPU paradigms, typical for fine-tuning workflows. We expect the qualitative conclusion to hold because multi-GPU systems are typically configured with more CPU cores, memory capacity, and I/O bandwidth. Under-provisioned nodes where aggregate GPU throughput outscales CPU resources could reintroduce data-path bottlenecks.
    \item \textbf{Startup overhead and memory constraints of pre-decoding.} This design choice introduces upfront wall-clock time to pre-decode the dataset before training starts, and constrains dataset size to the host system’s physical memory capacity. While these are not a problem for fine-tuning workflows, they prevent pre-training on internet-scale data.
    \item \textbf{Exploring GPU-based video decoding.} Our custom data engine relies entirely on CPU-bound video decoding. While we utilize the GPU for tensor normalization, we do not evaluate native, hardware-accelerated GPU decoding pipelines using \texttt{nvcodec}. However, any improvement in decoding is still bound by the same Amdahl envelope; we expect limited end-to-end gains unless the workload becomes data-bound.
\end{itemize}

%===============================================================================
%===============================================================================

\section{Conclusion}
This paper challenges the assumption that data loading and online augmentation act as primary computational bottlenecks in robot policy training. Through systematic profiling within the LeRobot ecosystem, we demonstrate that the data path accounts for 1.0--3.5\% of per-step training time, meaning even massive isolated data-infrastructure optimizations yield negligible end-to-end wall-clock improvements. Instead, severe training slowdowns stem primarily from format-specific serialization issues, like unindexed video decoding, rather than systemic data-loading architectures. These findings suggest that future robotics infrastructure efforts should shift focus away from lower-level data-loading abstractions and toward addressing actual upstream computational bottlenecks.

%===============================================================================
%===============================================================================

\clearpage
% The acknowledgments are automatically included only in the final and preprint versions of the paper.
\acknowledgments{We thank the members of the People, AI, and Robotics (PAIR) Lab for their valuable insights, helpful discussions, and ongoing support throughout this project. We would also like to thank the anonymous reviewers and the area chair for their constructive feedback.}

%===============================================================================

\bibliography{bibliography}  % .bib

%===============================================================================
% Appendix: must be in the same camera-ready PDF (CoRL 2026). Main text limit is
% 9 pages; acknowledgments, references, and appendix do not count toward it.
\newpage
\appendix
\section{Experimental Setup}
\label{app:setup}

\textbf{Hardware and System Specs.} The longer SmolVLA experiments were evaluated on a server-class L40S (48~GB), where SmolVLA training occupies $\sim$11.5~GB. All other experiments were run on a consumer PC with an RTX~4070~Ti~Super (16~GB) with 32~GB of system RAM.

\begin{table}[h]
  \centering
  \caption{System specs for both experimental configurations.}
  \label{tab:system_specs}
  \begin{tabular}{lcc}
    \toprule
     & Consumer PC & Server Node \\
    \midrule
    GPU & RTX~4070~Ti~Super & L40S \\
    GPU VRAM & 16~GB & 48~GB \\
    CPU cores & 32 & 32 \\
    Memory & 32~GB & 32~GB \\
    OS & Ubuntu 24.04.3 & RHEL9 \\
    Linux kernel & 6.17 & 5.14 \\
    \bottomrule
  \end{tabular}
\end{table}

\textbf{Datasets.} The ACT experiments use \texttt{lerobot/aloha\_static\_coffee}
($480\times640$, 4 cameras, 50 episodes, AV1, 50~fps) and
\texttt{mineral/lerobot\_mimicgen\_coffee\_d0} ($84\times84$, 2 cameras, 1000 episodes, AV1,
20~fps, with simulator evaluation); the SmolVLA experiment uses LIBERO-10~\citep{libero}.

\textbf{ACT policy.} ImageNet-pretrained ResNet-18 visual
  backbone; a transformer encoder--decoder with \texttt{dim\_model}=512, 8 heads,
  \texttt{dim\_feedforward}=3200, 4 encoder and 7 decoder layers; action chunk size
  100 with a single observation step; and the CVAE objective enabled
  (\texttt{latent\_dim}=32, 4 VAE-encoder layers, KL weight 10). We optimize for
  2{,}000 steps with learning rate $10^{-5}$ (backbone included), weight decay $10^{-4}$,
  gradient-norm clipping at 10, and seed 42; batch size is swept (8--64) per
  configuration. The 2{,}000-step budget is deliberate: these are timing-characterization
  runs rather than convergence runs, as per-step data-path and compute-path costs reach
  a stable steady-state regime within a few thousand optimizer steps, which is all our
  profiling requires. With $N$ cameras (4 for aloha, 2 for coffee) the visual-backbone
  forward is evaluated once per camera per step and is the dominant \emph{repeated}
  component of \texttt{updt\_s} -- the observation that motivates Section~\ref{sec:f6}. The smaller-ACT variant used at bs=16 in Section~\ref{sec:f1} reduces the transformer width and depth
  (\texttt{dim\_model}=256, 2 encoder / 4 decoder layers) to probe whether a
  cheaper model raises the data-path fraction.

\textbf{SmolVLA policy.} \texttt{HuggingFaceTB/SmolVLM2-500M-Video-Instruct} vision-language model (VLM) backbone with a frozen vision encoder, 16 VLM layers, and cross-attention enabled; action chunk size 50 with a single observation step; and state projection training enabled. We optimize over 100,000 training steps using AdamW with a peak learning rate of $10^{-4}$ decaying to $2.5\times10^{-6}$ over 30,000 steps via a cosine scheduler with a 1,000 step warmup. The optimization uses weight decay $10^{-10}$, $\beta=[0.9,0.95]$, gradient-norm clipping at 10, and seed 42; the batch size is fixed at 16. With 2 cameras (agent view and wrist view), input images are resized with padding to $512\times512$ resolution.

\newpage
\section{FFmpeg Fork Safety and Multiprocessing Architecture}
\label{app:fork_safety}

When scaling robot learning workloads, data-loading pipelines frequently rely on standard POSIX \texttt{fork()} semantics (the default multiprocessing start method in Linux-based PyTorch dataloaders) to spawn background workers. However, integrating high-performance video decoding libraries within a forked multi-process environment introduces severe concurrency risks, specifically concerning fork safety.

Our data engine utilizes the underlying FFmpeg library (via native C bindings) for
  video decoding. FFmpeg, alongside its internal thread pools and hardware-acceleration
  abstractions, is fundamentally not fork-safe. When a process calls \texttt{fork()},
  only the calling thread is duplicated in the child process; any internal mutexes,
  condition variables, or synchronization primitives held by other threads at the exact
  moment of the fork remain permanently locked in the child. If a background worker
  subsequently attempts to invoke an FFmpeg API call (such as allocating a decoder
  context or flushing a packet buffer), it encounters an unrecoverable deadlock. We
  observed precisely this class of failure during development: an early version of our
  pipeline guarded its episode cache with a read-write lock, and forked dataloader
  workers deadlocked when the lock was held across the \texttt{fork()} boundary. This
  firsthand experience with the fork-plus-held-lock hazard directly motivates the
  architecture described below, in which all FFmpeg decoding completes in the parent
  process and workers never touch the decoder.

To circumvent this limitation without sacrificing parallel data loading, our engine enforces a strict temporal separation between decoding operations and process spawning:
\begin{enumerate}
    \item \textbf{Synchronous Parent Preloading:} All video episodes in the dataset are entirely decoded and materialized into the immutable Apache Arrow intermediate representation (IR) within the parent process. This operation occurs sequentially before any parallel workers are initialized.
    \item \textbf{Context Disposal:} Once the Arrow buffers are fully populated, all active FFmpeg structures, file descriptors, and decoder handles are closed and cleanly disposed of in the parent process.
    \item \textbf{Worker Spawning:} The PyTorch training loop then forks the dataloading worker processes. Because the child processes only issue read requests against the static virtual memory mapped to the shared Arrow buffer, they never initialize or interact with the FFmpeg library.
\end{enumerate}

 \newpage
  \section{Feature Caching: Implementation and Analysis}
  \label{app:feature_cache}

  The dominant repeated cost in the ACT training step is the visual-backbone forward pass, re-evaluated over the same frames every epoch. Feature caching
  precomputes this output once and removes it from the training loop. Since Section~\ref{sec:f6} reports it as a preliminary result --- a
  1.3--1.5$\times$ training speedup with a faithful cache (a further 2.5$\times$ from pooled features), but 0/25 closed-loop success for both the cached
  policy and a stock ACT baseline --- we detail here how the cache is built and indexed, its relationship to the inference-time backbone, and what the
  rollout outcome does and does not tell us.

  \textbf{Construction and Indexing.} We run a frozen, ImageNet-pretrained ResNet-18 (\texttt{ResNet18\_Weights.IMAGENET1K\_V1}) in evaluation mode under
  \texttt{torch.no\_grad()}, resizing real-camera frames to $224\times224$ (the native $84\times84$ MimicGen frames are kept as-is) and ImageNet-normalizing
  them, then forwarding each once per \emph{(episode, frame, camera)} triple. We store one of two representations: a \emph{spatial-map} cache holding the
  pre-pooling \texttt{layer4} activation of shape $512\times H'\times W'$ ($512\times7\times7$ for $224$-resized inputs, $512\times3\times3$ for the
  $84\times84$ MimicGen frames), or a \emph{pooled} cache holding the global-average-pooled $512$-vector. Features are written in \texttt{float16} as one
  \texttt{.pt} shard per episode under a self-describing format (\texttt{data-core-feature-cache-v1}) with a manifest recording episodes, cameras, backbone,
  and shape. At training time the visual data path becomes an $O(1)$ lookup on the \emph{(episode, frame, camera)} key, returning the precomputed tensor to
  the action head; the backbone is never instantiated in the loop, which is what removes the repeated forward pass.

  \textbf{Map versus Pooled.} The representation dominates the result. On 40 aloha episodes the pooled cache is \textbf{47.5$\times$} smaller ($232.5$~MB vs.\
  $11.05$~GB), \textbf{2.5$\times$} faster to train ($24.6$ vs.\ $61.6$~s), and attains better held-out L1 ($0.0657$ vs.\ $0.0730$) --- both compressing the
  cache and regularizing the head on the demonstration distribution.
  
  \textbf{Equivalence to the Online Backbone.} The rollout failure is not a train/inference feature-skew artifact. The cache is an exact memoization of the same
  frozen ResNet-18 run online from live simulator images at rollout: identical weights, identical preprocessing, no backbone updates in either path. A
  same-head controlled comparison (cached vs.\ identical backbone run online) yields indistinguishable training L1 ($0.0623$ vs.\ $0.0624$), confirming the
  cache is faithful up to \texttt{float16} rounding.

  \textbf{Stock Baseline.} To separate caching from the training regime, we trained a stock LeRobot ACT (paper architecture with CVAE, trainable
  ResNet-18 backbone, same MimicGen coffee data) for 10k steps at batch size 64 and evaluated it, like the cached (spatial-map) policy, on 25 rollouts from
  deterministic initial states (400-step horizon). Training converged (L1 0.156$\to$0.074), yet the stock policy also scored \textbf{0/25}, matching the
  cached policy's \textbf{0/25}. The failure therefore stems from the short-budget behavior-cloning regime (hypotheses (1) and (4) below), not from caching.
  Establishing whether caching preserves task success requires a training budget at which the uncached baseline succeeds, which we leave to future work.

  \textbf{From Action Error to Task Success.} Candidate explanations, in decreasing order of likelihood given the stock baseline: \textbf{(1) open-loop vs.\ closed-loop} --- held-out L1 is a one-step error on demonstration states, whereas a rollout compounds small
  errors into unseen states where even faithful features are uninformative (the classic behavior-cloning failure, orthogonal to caching); \textbf{(2) loss of
  spatial grounding} --- the pooled cache discards the spatial layout precise manipulation likely needs, although the spatial-map cache evaluated above retains it and still fails;
  \textbf{(3) inference-semantics mismatch} --- the custom \texttt{FeatureACT} head omits the stock LeRobot ACT's CVAE, trainable backbone, and inference-time
  action normalization/temporal-ensembling; and \textbf{(4) limited training/coverage}. Because the stock baseline fails identically, (1) and (4) are
  sufficient to explain the result; the effect of (2) and (3) can be measured only once a longer training budget lifts the baseline above zero.

\newpage
\section{Supplementary Results}
\label{app:supp}

\begin{figure}[h!]
  \centering
  \includegraphics[width=0.5\linewidth]{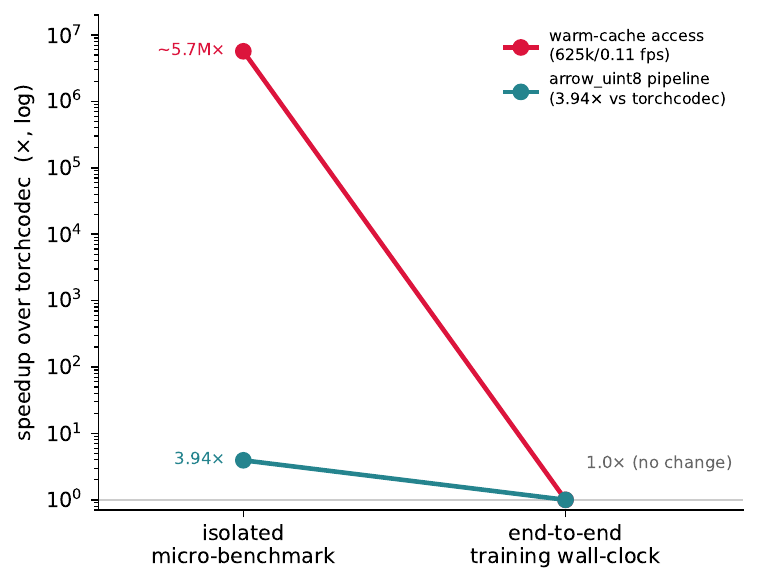}
  \caption{Isolated data-path speedups caused by warm-cache access
  ($\sim$$10^6\times$) and the \texttt{arrow\_uint8} pipeline (3.94$\times$) both
  collapse to no measurable speedup in end-to-end training wall-clock time.}
  \label{fig:f2}
\end{figure}

\begin{figure}[h!]
  \centering
  \includegraphics[width=0.7\linewidth]{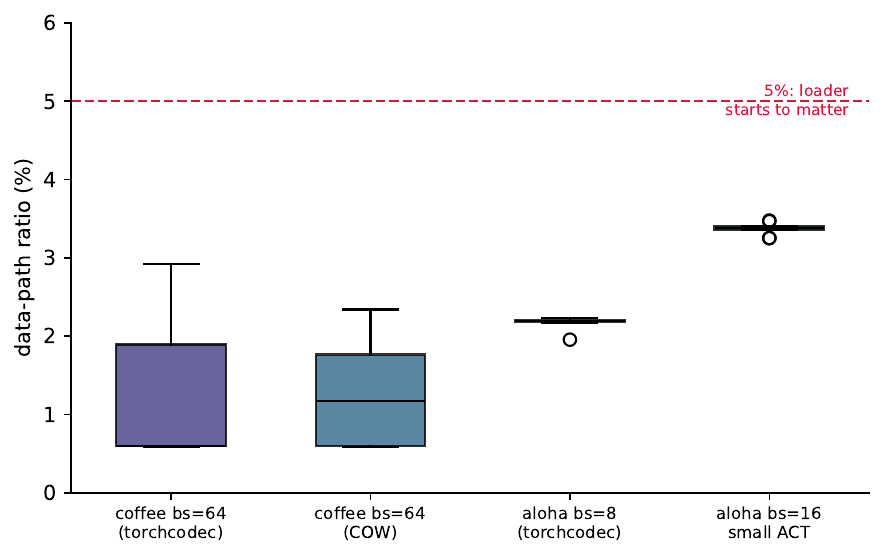}
  \caption{Distribution of the data-path ratio $r$ over 100-step windows for each configuration. Every configuration stays below the 5\% practical threshold (dashed line) discussed in Section~\ref{sec:f1}.}

  \label{fig:dist}
\end{figure}

\end{document}

%% file: preamble.tex
\usepackage[utf8]{inputenc} % allow utf-8 input
\usepackage[T1]{fontenc}    % use 8-bit T1 fonts
\usepackage{fonttable}
\usepackage[english]{babel} %for hyphenation rules
\usepackage[protrusion=true,expansion=true]{microtype}
\usepackage{lmodern}
\usepackage{comment}
\usepackage{cancel}
\usepackage{array}
\usepackage{colortbl}
\usepackage{csquotes}
\usepackage{listings}
\usepackage{nameref}
\usepackage{paralist}
\usepackage{xspace}
\usepackage{setspace}
\usepackage{array}
\usepackage{makeidx}
\usepackage{pdfpages}
\usepackage{times}

\usepackage{soul}

\usepackage{amsmath,amssymb} % define this before the line numbering.
\usepackage{amsfonts}       % blackboard math symbols
\usepackage{mathrsfs} %replacemnt for mathcal
\usepackage{mathtools}
\usepackage{array} %for table entries to be in center of cell
\usepackage{nicefrac}       % compact symbols for 1/2, etc.
\usepackage{textcomp}
\usepackage{bm} %bold symbols in math mode
\usepackage{pifont} % Provides \ding
\usepackage{graphicx}
\usepackage{wrapfig}
\usepackage{adjustbox} %add additional box operations in addition to resizebox
\usepackage{epsfig}
\pdfoutput=1
\graphicspath{{./figures/}}
\DeclareGraphicsExtensions{.pdf,.jpeg,.png,.jpg}

\usepackage{float}
\usepackage{placeins}
\usepackage{subcaption}
\usepackage[font=small,labelfont=bf]{caption}
\usepackage{lscape}      %Useful for wide tables or figures.

\usepackage{tikz}  %make graphs
\usepackage{makecell}
\usepackage{overpic}
\newsavebox{\subfigbox}

\usepackage{algorithm}
\usepackage[noend]{algpseudocode}

\usepackage{array} %for table entries to be in center of cell
\usepackage{tabularx}
\usepackage{multirow}
\usepackage{multicol}
\usepackage{booktabs}   % Publication quality tables
\usepackage{tablefootnote}
\usepackage{threeparttable}

\usepackage{paralist}
\usepackage{enumitem}
\setlist[itemize]{noitemsep,leftmargin=*,topsep=0in}
\setlist[enumerate]{noitemsep,leftmargin=*,topsep=0in}

\usepackage{url}            % simple URL typesetting
\usepackage[table]{xcolor}
\usepackage{color,colortbl} %superceded by xcolor
\usepackage{soul}

\PassOptionsToPackage{capitalize, noabbrev}{cleveref}

\usepackage{hyperref}
\usepackage{titlesec}
\titlespacing{\section}{0pt}{0.15\baselineskip}{0.1\baselineskip}
\titlespacing{\subsection}{0pt}{0.1\baselineskip}{0.08\baselineskip}
\titlespacing{\subsubsection}{0pt}{0.03\baselineskip}{0.02\baselineskip}
\renewcommand{\paragraph}[1]{\noindent\textbf{#1}\xspace}
\usepackage{blindtext}
\usepackage{bbm}

\usepackage{lipsum}

\definecolor{color1}{rgb}{0.6, 0.4, 0.05}
\definecolor{color2}{rgb}{0.0, 0.7, 0.7}
\definecolor{color3}{rgb}{0.35, 0.75, 0.0}
\definecolor{color4}{rgb}{0.4, 0.8, 0.0}
\definecolor{color5}{rgb}{0.5, 0.0, 0.5}
\definecolor{color6}{rgb}{0.99, 0.84, 0.69}
\definecolor{color7}{rgb}{0.8, 0.3, 0.3}
\definecolor{color8}{rgb}{0.2, 0.4, 0.8}
\definecolor{color9}{rgb}{0.6, 0.2, 0.6}
\definecolor{revision_color}{rgb}{1,0,0}